\documentclass[runningheads,a4paper]{llncs}

\usepackage{amssymb}
\usepackage{graphicx}

\usepackage{url}
\urldef{\mailsa}\path|{megumi}@i.kyoto-u.ac.jp|    
\newcommand{\keywords}[1]{\par\addvspace\baselineskip
	\noindent\keywordname\enspace\ignorespaces#1}

\usepackage{mathtools}
\newcommand{\argmin}{\mathop{\rm arg~min}\limits}

\def\Vec#1{\mbox{\boldmath $#1$}}

\begin{document}
	
\mainmatter  
	
\title{Sparse Elasticity Reconstruction and Clustering using Local Displacement Fields}
	
\titlerunning{Sparse Elasticity Reconstruction using Local Displacement}
	
%
%
\author{Megumi Nakao, Mitsuki Morita, Tetsuya Matsuda}
\authorrunning{Author's Name et. al}
	
\institute{Graduate School of Informatics, Kyoto University\\
\mailsa}
	
%
%
	
\toctitle{Lecture Notes in Computer Science}
\tocauthor{Authors' Instructions}
\maketitle

\begin{abstract}
This paper introduces an elasticity reconstruction method based on local displacement observations of elastic bodies. Sparse reconstruction theory is applied to formulate the underdetermined inverse problems of elasticity reconstruction including unobserved areas. An online local clustering scheme called a {\it superelement} is proposed to reduce the number of dimensions of the optimization parameters. Alternating the optimization of element boundaries and elasticity parameters enables the elasticity distribution to be estimated with a higher spatial resolution. The simulation experiments show that elasticity distribution is reconstructed based on observations of approximately 10\% of the total body. The estimation error was improved when considering the sparseness of the elasticity distribution.
		
\end{abstract}

\section{Introduction}
Tissue elasticity can be an indicator for the detection or diagnosis of a lesion in an organ. A variety of elastography techniques such as magnetic resonance elastography (MRE) \cite{Mariappan2010} and ultrasound elastography \cite{Shiina2013} have been developed to measure in vivo elasticity information. In addition to elasticity imaging, model-based estimation methods \cite{Doyley2012}\cite{Goksel2013}\cite{Morita2017} for reconstructing elasticity mathematically have been investigated. Elastic modulus of tissue was estimated based on the Navier-Stokes equation by solving an optimization problem using a finite element (FE) model. Its mesh adaptation was also investigated to improve the accuracy of tissue-elasticity reconstruction \cite{Goksel2013}. Although these methods show that elasticity of an observable area can be spatially identified, the whole shape and displacement are needed for elasticity reconstruction. However, application of the model-based approach has been restricted because the entire shape of organs cannot be obtained in many clinical situations such as ultrasound and intraoperative imaging. Recent study reports external force can be estimated using locally observed displacements of the deformed state \cite{Sakata2017}. To the best of our knowledge, without our preliminary study \cite{Morita2017}, no report has appeared investigating elasticity reconstruction that includes unobservable areas of elastic bodies. 
	
This paper introduces an elasticity reconstruction method of elastic bodies using local displacement fields. We focus on cases involving hard inclusion that cannot be observed within the estimated target, and propose a method to estimate the spatial distribution of elasticity using only local or surface deformation. Sparse reconstruction theory \cite{Candes2008} is applied to formulate this underdetermined inverse problem, and the number of deformation patterns in the elastic body model are used to improve estimation accuracy. We extend the concept of superpixels \cite{Achanta2012}\cite{Wu2017} to FE mesh models, and investigate whether the {\it superelement} framework can be adapted to models with high spatial resolution. This concept has a potential for a variety of clinical application\cite{Nakao2010}\cite{Nakao2014}. By estimating elasticity based on locally observed data, the areas of elastgraphy can be further extended. If the elasticity of organs is reconstructed from surface deformation, intraoperative guide and vision-based tumor localization will be possible without additional hardware setup.

\section{Elasticity Reconstruction Using Local Displacements}
	
\subsection{Problem definition}
Fig. 1 shows an outline of the proposed elasticity reconstruction framework. We assume that the model shape is imaged using computed tomography (CT) or magnetic resonance imaging (MRI) and assumed to be self-evident. The goal is to output the elasticity distribution of the elastic bodies, including areas that cannot be observed. We specifically focus on a situation in which hard inclusion is located in the unobservable area. The displacement fields of the elastic body that can be locally observed are used as the input. For instance, a vision-based approach that obtains feature-based tracking \cite{Zhao2016} and ultrasound imaging \cite{Shiina2013} are available to measure the displacement of sampled points. Therefore, we suppose that the movement of each visible point on the elastic body surface or the internal structure is available. In addition, when solving FEM, we assume that external forces contributing to organ deformation is known and other small forces between neighboring tissues are neglected. Many researchers report that the external force can be measured using a force sensor in the forceps. Also, this condition does not lose generality in conventional elasticity imaging because external forces or loads are explicitly given by transducer of a probe.
	
To design the optimization process, calculation time and stability of convergence that arise with the increasing number of elasticity parameters should be considered. To address these issues, the superpixel concept \cite{Achanta2012} is applied to the mesh model. In image processing, pixels with similar pixel information are considered a single area called a superpixel. In this study, pixels are replaced with the elements comprising a mesh model, the pixel information is replaced by elasticity, and the area comprising a group of elements with similar elasticities is referred to as a superelement. In other words, the elasticity parameters are expressed in superelement units. 
	
\begin{figure}[t]
	\begin{center}f
		\includegraphics[width=10.5cm]{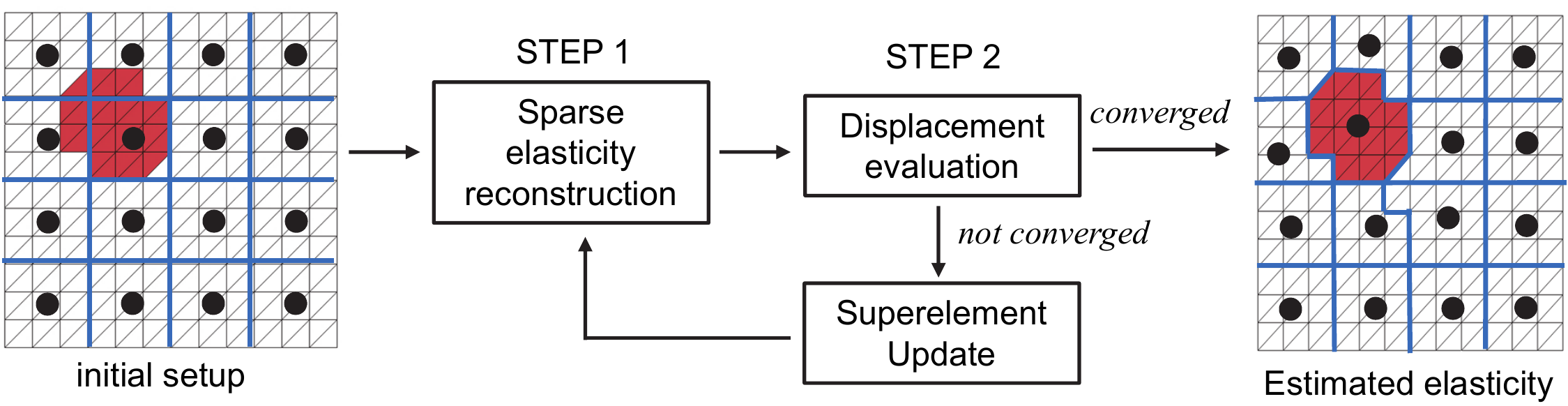}
	\end{center}
	\vspace{-0.5cm}
	\caption{An outline of the proposed elasticity reconstruction framework.}
\end{figure}
		
\subsection{Mathematical formulation}

In the proposed framework, a linear FE model is used to formulate the problem. Although nonlinear modeling is required to handle large deformation, small deformation (e.g. within 10mm) is sufficient for the proposed elasticity reconstruction, and applying the linear model does not lose generality in this research context. According to the equation used in linear finite element analysis, we obtain $\Vec{u} = K(\Vec{E})^{-1} \Vec{f} = L(\Vec{E}) \Vec{f}$. Here, $\Vec{f}$ is the force vector applied to each vertex, $\Vec{u}$ is the displacement, $\Vec{E}$ is the stiffness parameter (e.g., Young's modulus and Poisson's ratio) vector, and $K$ is the stiffness matrix. By categorizing all the vertices of a mesh model as observable $\Vec{v_o}$ and unobservable $\Vec{v_i}$ vertices, this equation can be rewritten as 
	
	\begin{eqnarray}
	\left[ 
	\begin{array}{c}
	\Vec{u}_o \\
	\Vec{u}_i \\
	\end{array} 
	\right]
	=\left[
	\begin{array}{cc}
	L_{oo} & L_{oi} \\
	L_{io} & L_{ii} \\
	\end{array}
	\right]
	\left[
	\begin{array}{c}
	\Vec{f}_o \\
	\Vec{f}_i \\
	\end{array}
	\right]
	\end{eqnarray}
	
The observed displacement $\Vec{u_o}$ is expressed as $\Vec{u}_o=L_{o} \Vec{f}$ using $L_{o} = \left[ L_{oo} \hspace{0.1cm} L_{oi} \right] $. To consider multiple patterns of the deformation state, using the locally observed displacements matrix $U_o= \left[\Vec{u}_{1o} \hspace{0.1cm} \Vec{u}_{2o} \hspace{0.1cm} \Vec{u}_{3o} \cdots \right]$, the minimization problem for elasticity reconstruction can be expressed as 
	
\begin{eqnarray}
\Vec{E}^*= \argmin_{\scalebox{0.6}{\Vec{E}}}\lVert U_o- {U'} _o\rVert_F.
\label{optmiseUo}
\end{eqnarray}
We note that this equation is underdetermined because the number of observed vertices are smaller than the dimension of stiffness vector $\Vec{E}$. To make this minimization problem solvable, the concepts of superelement-based alternating optimization and sparseness of tissue elasticity are utilized in this paper.
	
\subsection{Alternating optimization of element boundaries and elasticities}
The $superelement$ concept locally clusters elements with homogeneous elasticity. Using this framework, the number of parameters to be optimized can be reduced even when the shape of the model is complex, which reduces the computation time and improves the stability of the optimization problem. 
	
Each superelement has two attributes: elasticity $E_i$ and central coordinate $C_i$, and serves as a medium through which elasticity values propagate to the member elements of the mesh model. Here, the goal is to optimize $(E_i, C_i)$. To improve the stability of the optimization, we introduce simple constraints on the central coordinates of the superelements into the optimization framework as follows. 
	
\begin{eqnarray}
\Vec{E}^* = \argmin_{\scalebox{0.6}{\Vec{E}}}\lVert \Vec{U_o}- \Vec{U'_o}\rVert_F+\omega\lVert {\Vec{C}}- \Vec{C_0}\rVert_F
\label{optmiseUo2}
\end{eqnarray}
Here, $\omega$ is a weight parameter. This constraint is intended to prevent the superelement central coordinates from becoming dispersed, and the penalty is applied based on the moving distance from the initial position. 

As a initial setup, the central coordinates of the superelement are arranged equidistantly within the model, and the superelement is placed into the initial state. The element boundary and elasticity are updated alternately as follows:
		
\begin{description}
\item[STEP 1 Optimization of the superelement elasticity]\mbox{}\\
The elasticity parameters $\Vec{E}$ of the superelements are first optimized while the central coordinates are fixed. The elasticity of superelements is propagated to the member elements, and the model deformation is obtained using Eq. (1). The stiffness parameter $\Vec{E}$ is iteratively updated until the evaluation value in Eq. (3) is converged.

\item[STEP 2 Optimization of the central coordinates]\mbox{}\\
The central coordinates $\Vec{C}$ are then updated based on the optimized value of the elasticity parameters $\Vec{E}$. When the central coordinates of the superelement are updated, the member tetrahedral elements that belong to each superelement is also updated. This update scheme is similar to the superpixel framework\cite{Achanta2012}. After the central coordinates are converged, STEP 1 is restarted using the new element boundaries. 
\end{description}

\subsection{Sparse elasticity reconstruction}
To improve the estimation accuracy, we focus on the sparseness of tissue elasticity. Sparseness is mathematically defined as a state where only a small number of non-zero values exist within the whole data. It is known that tumors and lesions are harder than healthy tissues; however, lesions are localized within tissues as a whole. If we assume that the elasticity in areas other than the lesion site is uniform, we can consider that, in the lesion, the elasticity locally deviates from the expected distribution of the organ elasticity. Considering the sparseness of the elasticity gradient, it is possible to add constraints to the elasticity such that the major part of an elastic body is uniform. The optimization problem with respect to the sparseness of the elasticity gradient can be formulated as follows:
	
\begin{eqnarray}
\Vec{E}^*= \argmin_{\scalebox{0.6}{\Vec{E}}} \lVert U_o- {U'} _o\rVert_F+\omega\lVert {C}- {C_0}\rVert_F+\lambda \lVert \Vec{E} - \Vec{E_0} \rVert_1,
\label{optmiselamda}
\end{eqnarray}
where the elements of \Vec{E} are the elasticity parameters to be solved, the elements of $\Vec{E_0}$ are the reference elasticity values for healthy tissue, and $\lambda$ is the coefficient controlling the sparseness of the elasticity gradient. The greater the value of $\lambda$, the more restricted the localized lesion area, causing the estimation results for the elasticity distribution to approach uniformity.

\section{Experiments and Results}
	
\subsection{Performance of sparse elasticity reconstruction}
In the simulation experiments, we used a simple plate model with 98 vertices and 216 tetrahedral elements. As shown in Fig. 2(a), 14 fixed points (red) were configured. The plate model was partitioned into $6\times6$ equal regions, and the centers of the superelements were positioned in the center of each region. In the first experiment, the center of the superelement was fixed to simplify the problem setting, and therefore the estimation target was a 36-dimensional Young's modulus $\Vec{E}$. The objective function in Eq. (3) is solved using CMA-ES. The Young's modulus of the soft regions was set to 35.8 kPa, and $2\times2$ hard regions were set to 117.6 kPa. The initial values of all $\Vec{E}$ elements were set to 35.8 kPa. This was based on measurements of the elasticity of an actual 3D printer plate. Although Poisson's ratio can be computed in the proposed framework, 0.4 was uniformly assigned to simplify the experiments. 

Regarding observation conditions, the number of observed vertices was increased by five sequentially along the -x axis on the upper surface. We assumed that an external force was exerted on the surface of an organ by forceps or by the transducer of a probe. Therefore, the magnitude and direction of the external force are regarded to be uniform over all neighboring vertices. At the contact points, a force of 9.8 N was directed along the +y axis, the -y axis, and the +x axis (Fig. 2b). Then, one, two, or three pattern deformations were prepared.

The estimation error of elasticity reconstruction results are shown in (c). The horizontal axis is the number of observed vertices, and the vertical axis is the root-mean-squared error (RMSE) values of the elasticity estimation results. To evaluate the performance of sparse reconstruction, the results include the reference case by setting zero to the sparseness coefficient $\lambda$. When the relationship between the observed vertices and estimation accuracy was determined based on the deformation state of one pattern, the RMSE was 431.4 kPa for five observed vertices and 6.0 kPa for 20 observed vertices. This finding confirms that the estimation accuracy increases with the number of observed vertices. By considering sparseness of elasticity gradient, it is possible to precisely estimate the elasticity by observing vertices in approximately 20\% of the whole body.

\begin{figure}[t]
	\begin{center}
		\includegraphics[width=12.0cm]{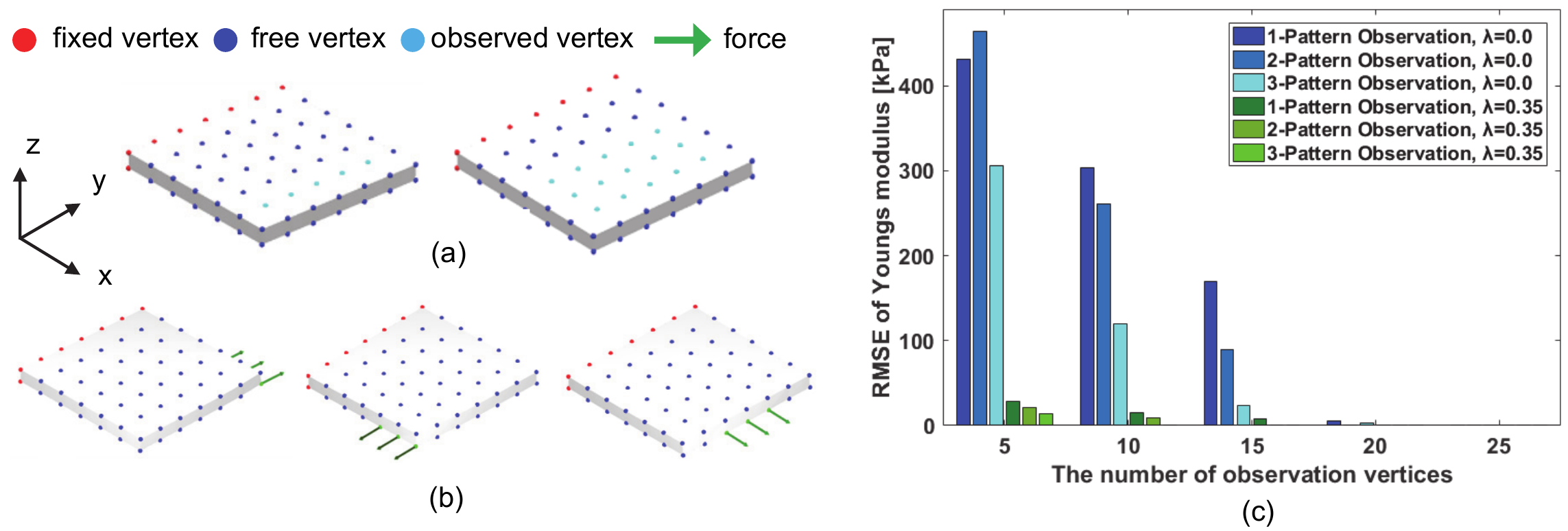}
	\end{center}
	\vspace{-0.75cm}
	\caption{ Performance of elasticity reconstruction for a plate model. (a) observation conditions (5 and 15 observed vertices for example), (b) three patterns of external forces and (c) RMSE of estimation results.}
\end{figure}	

\begin{figure}[t]
	\begin{center}
		\includegraphics[width=10.5cm]{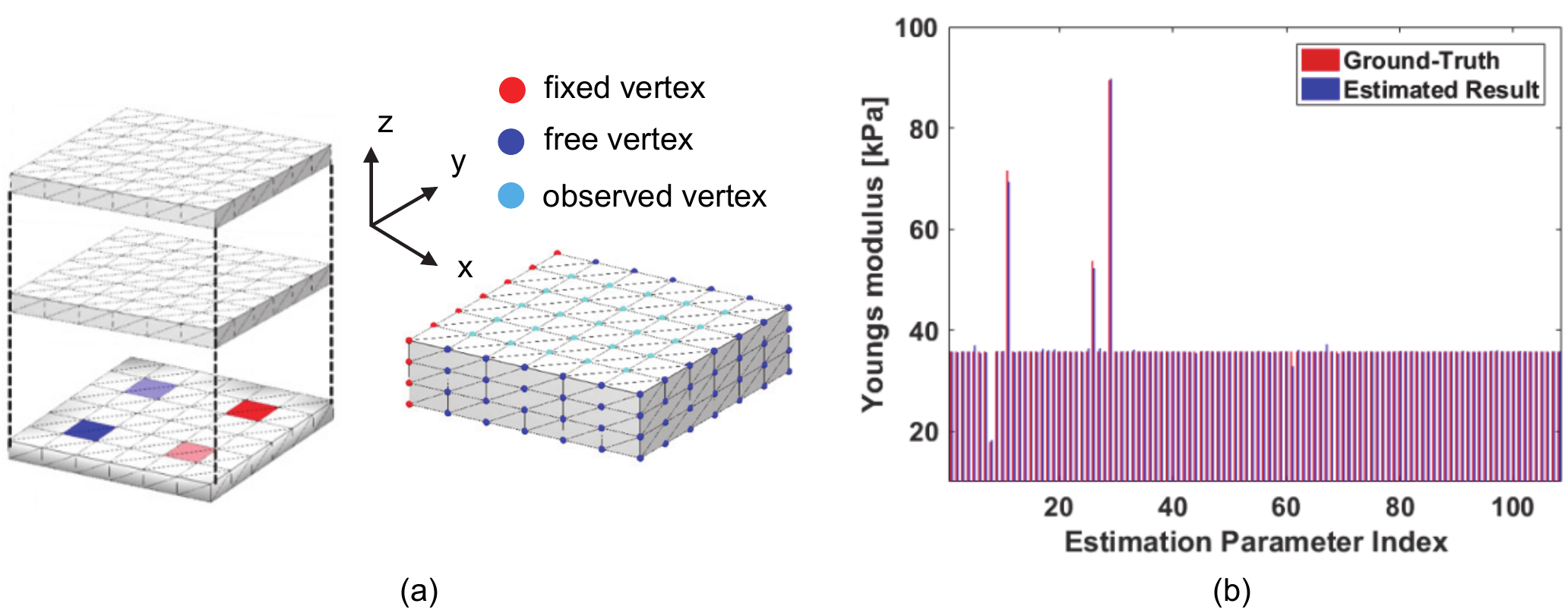}
	\end{center}
	\vspace{-0.5cm}
	\caption{Elasticity reconstruction results of a three-dimensional plate model with plural stiffness, (a) Ground-truth (blue: 17.9kPa, light blue: 53.6kPa, light red: 71.5kPa and red: 89.4kPa) and (b) the estimation result.}
\end{figure}	

Additionally, in terms of the relationship between the number of observed elastic body deformation state patterns and the estimation accuracy, when 15 vertices were observed, the RMSE was 169.3 kPa with one pattern observation, 90.0 kPa with two pattern observations, and 23.2 kPa with three pattern observations. This finding confirms that the estimation accuracy improves when more observations are conducted. These results also show that when the elasticity distribution is estimated using local observations, it is better to observe a variety of deformation states to achieve a more precise estimate
	
We also examined the elastic modulus estimation for a three-dimensional plate model shown in Fig. 3 with 196 vertices and 648 tetrahedral elements. This FE model was divided into $6\times6\times3$ rectangular regions that represented distribution of Young's moduli. We observed 25 points on the upper surface of the elastic model. An external force of 9.8 N was applied to three vertices of the lower surface of the model, similar to the previous experiment. The results shown in Fig. 3 have an RMSE of 0.45 kPa. The index of the Young's modulus parameter was set in ascending order of the x, y, and z axes. These results suggested that the proposed method can be applied to an elastic model with plural stiffness.

\subsection{Elasticity reconstruction with superelement optimization}
The purpose of the next experiment is to apply the proposed method to a model with a higher spatial resolution than provided by the superelements. The relationship between the superelement's initial allocation and estimated accuracy was investigated, and the estimation performance of the proposed alternating optimization using the nine superelements was confirmed. We carried out an elasticity estimation experiment wherein the superelements were updated. The initial allocation of the superelement centers, as shown in Fig. 4(a), are set in four ways. The other experimental conditions are the same as those in Fig. 3.

The RMSEs for each result were (a) 8.5 kPa, (b) 5.8 kPa, (c) 7.7 kPa, and (d) 3.1 kPa for each of the four initial superelement center allocation methods. The estimation accuracy for (d), which had the largest number of superelements, was the highest, whereas it was the worst for (a), which had the smallest number of superelements. Additionally, we confirmed that when we compare (b) and (c), even when the number of superelements is the same, a difference occurs in the estimation margin of error based on the initial allocation state. These results show that the estimation accuracy is influenced by the number of superelements and initial allocation state of their centers.
	
\begin{figure}[t]
	\begin{center}
		\includegraphics[width=11.0cm]{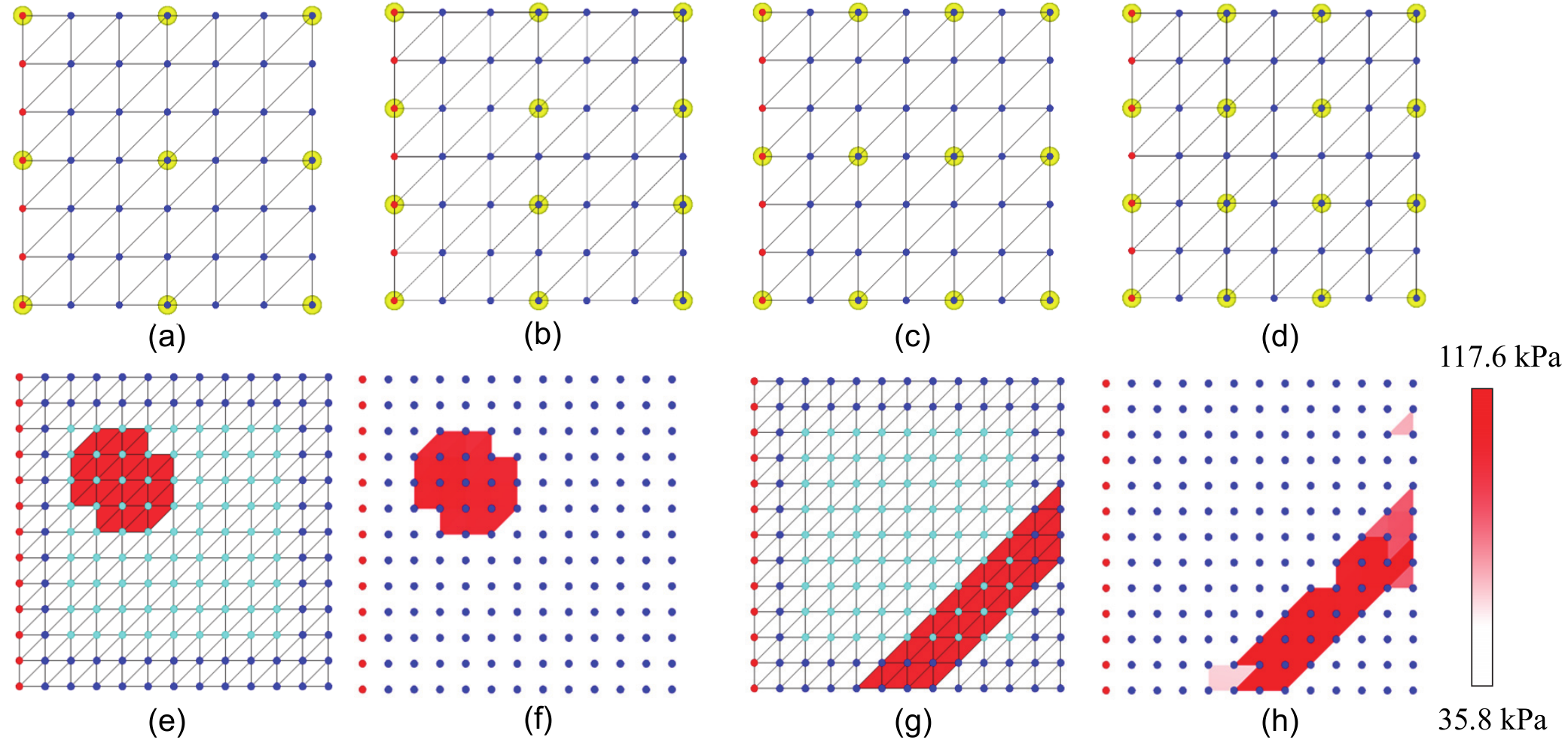}
	\end{center}
	\vspace{-0.5cm}
	\caption{Elasticity reconstruction results with superelement optimization.}
\end{figure}	
	
In the last experiment,	we investigated whether the proposed method can be applied to models with high spatial distribution. The experimental conditions for the model used to estimate the elasticity are shown in Fig. 4(e) and (g). The model shape were the same as the plate model used in Fig. 3. However, the mesh model comprised 338 vertices and 864 tetrahedral elements. There were 26 fixed points, 5 contact points, and 81 observation points set on the top surface. When estimating this model without using the superelement concept, 864-dimensional parameters must be optimized, and it is difficult to calculate this estimation within a realistic time. In this experiment, $6 \times 6$ superelements were placed in the model, so the number of superelements was significantly lower than that of tetrahedral elements. Using the proposed method, the problem of reconfiguring elasticity for 864 elements is estimated as a problem of optimizing 36 dimensions. The results of estimating elasticity for the two elastic models shown in Fig. 4(e) and (g) are shown in Figs (f) and (h), respectively. In Fig. 4(f), the result of this was that, based on an observation of approximately 24\% of the total, estimation was achieved with RMSE of 1.2 kPa and maximum error of 7.0 kPa. In contrast, as shown in Fig. 4(h), the approximate position of the hard section can be identified, but the accuracy of estimating the elasticity was lower. The elements demonstrating a maximum error up to 90.8 kPa were observed. This is thought to be influenced by the fact that each element is correct and not classified into superelements. Based on the limitation we have obtained, to develop ideas on more efficient objective function is future work. 

\section{Conclusions}
This paper proposed a sparse elasticity reconstruction method using the locally observed displacements of elastic bodies. The sparse modeling approach and the superelement-based alternating optimization scheme were introduced to improve estimation performance and optimization stability. Future work includes application to an organ-shaped model and improvement of the algorithms.

\end{document}